# Bridging Brain Connectomes and Clinical Reports for Early Alzheimer's Disease Diagnosis

Jing Zhang[1] Xiaowei Yu[1] Minheng Chen[1] Lu Zhang[2] Tong Chen[1] Yan Zhuang[1] Chao Cao[1] Yanjun Lyu[1] Li Su[3,4] Tianming Liu[5] Dajiang Zhu[1]

[1]Computer Science and Engineering, The University of Texas at Arlington, Arlington, TX, USA
{jxz7537, xxy1302, mxc2442, txc5603, yxz8653, cxc0366, yxl9168}@mavs.uta.edu, dajiang.zhu@uta.edu

[2]Department of Computer Science, Indiana University Indianapolis, IN, USA
lz50@iu.edu

[3]Department of Psychiatry, School of Clinical Medicine, University of Cambridge, UK

[4]Neuroscience Institute, School of Medicine and Population Health, University of Sheffield, UK
l.su@sheffield.ac.uk

[5]School of Computing, The University of Georgia, Athens, GA
tliu@uga.edu

**Abstract.** Integrating brain imaging data with clinical reports offers a valuable opportunity to leverage complementary multimodal information for more effective and timely diagnosis in practical clinical settings. This approach has gained significant attention in brain disorder research, yet a key challenge remains: how to effectively link objective imaging data with subjective text-based reports, such as doctors' notes. In this work, we propose a novel framework that aligns brain connectomes with clinical reports in a shared cross-modal latent space at both the subject and connectome levels, thereby enhancing representation learning. The key innovation of our approach is that we treat brain subnetworks as tokens of imaging data, rather than raw image patches, to align with word tokens in clinical reports. This enables a more efficient identification of system-level associations between neuroimaging findings and clinical observations, which is critical since brain disorders often manifest as network-level abnormalities rather than isolated regional alterations. We applied our method to mild cognitive impairment (MCI) using the ADNI dataset. Our approach not only achieves state-of-the-art predictive performance but also identifies clinically meaningful connectome-text pairs, offering new insights into the early mechanisms of Alzheimer's disease and supporting the development of clinically useful multimodal biomarkers.

**Keywords:** Medical Vision-Language Model, Multimodal Learning, MCI.

## 1 Introduction

Alzheimer's disease (AD) is one of the most prevalent neurological disorders in the United States, progressively impairing social, linguistic, and cognitive functions. With no known cure, early diagnosis – particularly at the mild cognitive impairment (MCI) stage, a precursor of AD – is critical for timely intervention and treatment planning [1]. Current diagnostic approaches rely on clinical evaluations, cognitive assessments, and neuroimaging techniques. However, early diagnosis remains challenging due to the heterogeneity in disease manifestation and pathology. Overcoming this challenge requires a comprehensive analysis of both objective imaging data and subjective clinical measures, such as doctors' notes and other textual reports. Consequently, there is an urgent need for models that can effectively integrate multimodal clinical data, including neuroimaging and text-based information, a research direction that has gained significant attention in AD studies.

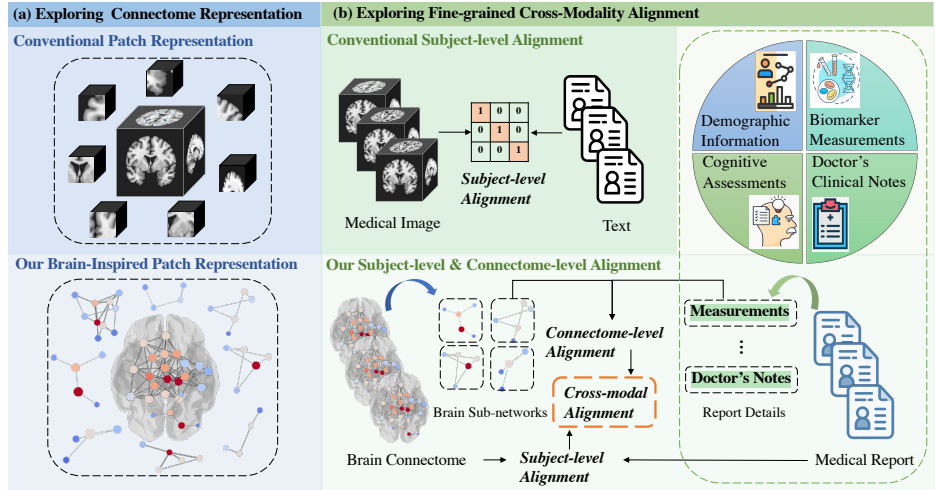

**Fig. 1.** Limitations of previous methods: (a) insufficient brain connectome understanding and (b) limited fine-scale cross-modal representation, and our corresponding improvement.

Recent advancements in vision-language models (VLMs) have demonstrated the potential of multimodal learning in brain healthcare. For instance, Xue et al. [2] utilized multimodal data, including both imaging and comprehensive non-imaging information, for differential dementia diagnosis. Similarly, Zhang et al. [3] proposed Brain-Adapter, a learnable embedding adapter designed to align multimodal data from imaging and non-imaging sources. However, current methods largely adopt strategies designed for general image processing—such as treating image patches as tokens in Euclidean space (Fig. 1a) - without considering the complex neurological and pathological information inherent in brain images. As a result, these methods struggle to extract meaningful features that can effectively align heterogeneous image-derived abnormalities with individual clinical reports in real-world settings. Additionally, beyond predictive accuracy, an ideal model should also provide interpretability, illustrating how multimodal data contribute to the overall diagnosis.

To address these challenges, we propose a novel framework that aligns brain connectomes with clinical reports in a shared cross-modal latent space at both the subject and connectome levels, to enhance representation learning (Fig. 1b). The key innovation of our approach is that we treat brain subnetworks as tokens of imaging data, rather than raw image patches, to align with word tokens in clinical reports. This design enables a more efficient identification of system-level associations between neuroimaging findings and clinical observations, which is critical since brain disorders often manifest as network-level abnormalities rather than isolated regional alterations. We applied our method to MCI diagnosis using the ADNI dataset, achieving state-of-the-art predictive performance while also identifying clinically meaningful connectome-text pairs. These findings provide new insights into the early mechanisms of Alzheimer's disease and support the development of clinically useful multimodal biomarkers.

## 2  Methodology

As illustrated in Fig. 2, our proposed framework includes two major parts: (1) Multi-modality Representation and (2) Cross-Modality Alignment. The first component focuses on efficient encoding of brain connectome and clinical reports separately, the second part aims to capture the complex relationship between image and non-image derived encodings via a novel cross-modality alignment approach.

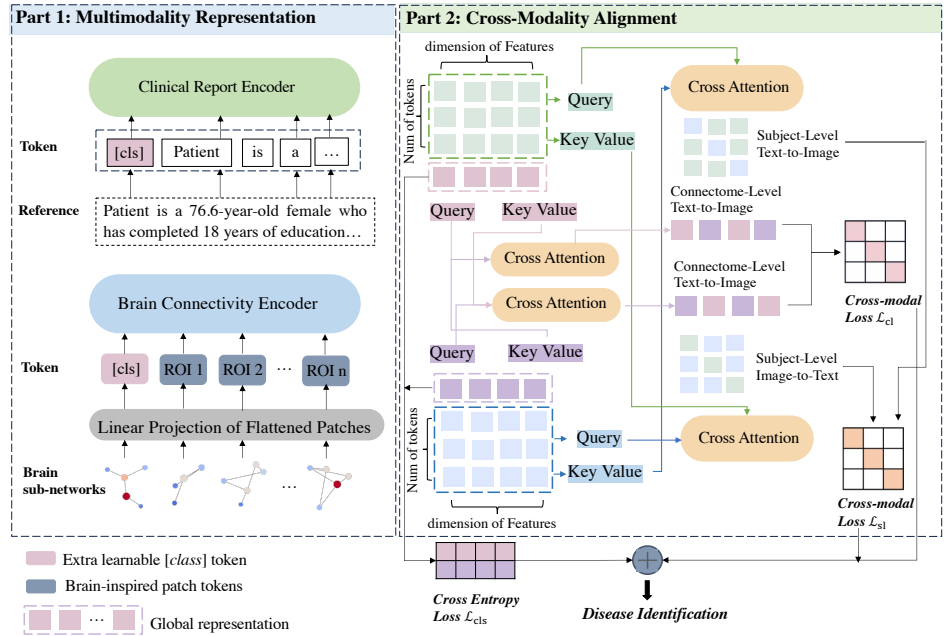

**Fig. 2.** Overview of the proposed framework. (1) Multimodality Representation, responsible for extracting brain connectome and clinical records representations, and (2) Cross-Modality Alignment, including subject-level alignment and connectome-level alignment.

### 2.1 Multimodality representation

**Brain Connectome Representation.** In this section, we introduce a novel brain-inspired patch representation to guide the encoder understand complex brain connectome. For a given subject, the SC matrix $SC \in R^{N \times N}$ represents the whole brain connectivity, where $SC_{i,j} \in R$ is the number of fibers connecting brain regions $i$ and $j$. To capture localized connectivity patterns, we define brain sub-network as patches $SC_{patch}^n \in R^N$. Here, each sub-network is defined as the structural connectivity between one brain region to all the other brain regions, served as a connectome-level representation vector. Next, we employed a linear fully connected layer as the patch embedding layer to process SC patches as $SC_{patch} \in R^{N \times D} = SC_{patch} \in R^{N \times N} \times W^T + b$, where $N$ is the number of brain sub-networks and $D$ corresponds to the feature dimension for each sub-network, $W^T$ and $b$ are the weight and bias of the linear layer, respectively. To aggregate information from the whole-brain network, we introduce a learnable class token $SC_{cls} \in R^D$, which enables the model to combine information from all brain sub-networks into a comprehensive connectome-level representation. This class token is concatenated with all brain sub-network tokens as $SC_{embedding}$.

Lastly, the generated $SC_{embedding} \in R^{(N+1) \times D}$ is fed into a brain connectivity encoder which adopts the vanilla ViT as its backbone. Our encoder contains a series of transformer encoder layers, each consists of two Layer Normalization (LN) blocks, one Multi-Head Self-Attention (MSA) block, and one Multi-Layer Perceptron (MLP) block: $o_i' = MSA(LN(o_{i-1})) + o_{i-1}; o_i = MLP(LN(o_i')) + o_i'$, where $o_{i-1}$ represents the input to the $i-th$ transformer layer, $o_i'$ and $o_i$ denote the intermediate and final outputs of the layer. Through this sequential processing, the brain connectomes are transformed into a local representation $X_{local} \in R^{N \times D}$ capturing sub-network features and a global representation $X_{global} \in R^{1 \times D}$ summarizing the entire brain features. This process is illustrated in the bottom of Fig. 2, part 1.

**Clinical Reports Representation.** To process and understand the semantic content of clinical reports, a text encoder is employed, as shown in the top section of part 1 in Fig.2. In this work, we adopt BERT as the backbone of our clinical report encoder for semantic modeling. Specifically, our clinical report encoder transforms each input report $T$ into a local token-level representation $V_{local} \in R^{M \times D}$ capturing token-specific features, and a global report-level representation $V_{global} \in R^{1 \times D}$ summarizing the overall semantic content. Here, $M$ is the number of tokens in each subject's reports and $D$ denotes the feature dimension.

### 2.2 Multi-modality alignment

**Connectome-level Alignment.** To explore meaningful associations between brain sub-networks and clinical report tokens, we propose a fine-grained connectome-level alignment module: the brain sub-network embeddings $X_{local}$ are used as query, while clinical token embeddings $V_{local}$ serve as the key and value in a cross-attention mechanism to capture the relationships between each brain sub-network and text tokens, producing the attention output $brain2text_{cl} = \text{softmax}\left(\frac{X_{local} \cdot V_{local}^T}{\sqrt{D}}\right) \cdot V_{local}$. Similarly, we compute the reverse attention flow where $V_{local}$ serve as query and $X_{local}$ serve as key and value to generate $text2brain_{cl} = \text{softmax}\left(\frac{V_{local} \cdot X_{local}^T}{\sqrt{D}}\right) \cdot X_{local}$, enabling bidirectional information exchange across modalities. The attention outputs are then normalized and used to compute cosine similarity matrix $S_{cl} \in R^{N \times M}$ that quantifies the connectome-level alignment between each brain sub-network and clinical token.

For each brain sub-network $i$, we maximize its alignment with the relevant clinical tokens through weighted similarity: $\mathcal{L}_{cl}^{b \to t} = -log(\sum(softmax(S_i) \cdot S_i)$, where $S_i$ represents the similarity scores between brain sub-network $i$ and all tokens. Similarly, for each token $j$, we maximize its alignment relevant brain sub-networks through: $\mathcal{L}_{cl}^{t \to b} = -log(\sum(softmax(S_j) \cdot S_j)$, where $S_j$ represents the similarity scores between clinical token $j$ and all sub-networks. The connectome-level alignment is optimized by the combined loss $\mathcal{L}_{cl} = 1/2\ (\mathcal{L}_{cl}^{b \to t} + \mathcal{L}_{cl}^{t \to b})$.

**Subject-Level Alignment.** To establish a holistic alignment across different patients' brain connectivity patterns and their clinical diagnostic reports, we implement a subject-level alignment module: we utilize the brain connectivity representations $X_{global}$ and semantic representation $V_{global}$ of clinical report to compute bidirectional cross-attention features. The subject-level brain-to-text attention produces $brain2text_{sl}$ by treating $X_{global}$ as the query and $V_{global}$ as key and value pairs. Similarly, text-to-brain attention generates $text2brain_{sl}$ using $V_{global}$ as query and $X_{global}$ as key and value. The cross-modal features are L2-normalized, and a cosine similarity matrix is computed as $S_{sl}^{i,j} = \frac{brain2text_{sl}^i \cdot text2brain_{sl}^j}{\|brain2text_{sl}^i\| \|text2brain_{sl}^j\|}$, where $S_{sl}^{i,j}$ quantifies the similarity between the $i$-th patient's brain connectivity representation and the $j$-th patient's text representation. The subject-level alignment is optimized by the InfoCNE loss to ensures that the paired brain connectivity and clinical report embeddings are close in the shared latent space:

$$\mathcal{L}_{sl}^{b \to t} = -\frac{1}{N}\sum_{i=1}^{N} log\ \frac{exp(S_{sl}^{i,j}/\tau)}{\sum_{k=1}^{N} exp(S_{sl}^{i,k}/\tau)}, \mathcal{L}_{sl}^{t \to b} = -\frac{1}{N}\sum_{i=1}^{N} log\ \frac{exp(S_{sl}^{i,j}/\tau)}{\sum_{k=1}^{N} exp(S_{sl}^{k,i}/\tau)} \quad (1)$$

The overall subject-level alignment can be denoted as: $\mathcal{L}_{sl} = 1/2\ (\mathcal{L}_{sl}^{b \to t} + \mathcal{L}_{sl}^{t \to b})$.

**Overall objective.** To achieve the downstream task of MCI/NC prediction, $X_{global}$ and $V_{global}$ were combined and passed through a fully-connected layer with a nonlinear activation function. A balancing strategy was applied to the cross-entropy loss $\mathcal{L}_{cls}$. We trained our model by jointly optimizing the loss: $\mathcal{L} = \mathcal{L}_{cl} + \mathcal{L}_{sl} + \mathcal{L}_{cls}$.

## 3 Experiments and results

### 3.1 Data Preprocessing and Experimental Design

This study utilized both imaging and non-imaging data from ADNI. The imaging dataset consists of T1-weighted structural MRI and DTI data from 418 subjects including 301 normal controls (NC) (183 females, 70.25±5.91Y; 118 males, 71.37±5.92Y) and 117 individuals with MCI (43 females, 70.17±7.32Y; 74 males, 73.21±6.86Y). Standard imaging preprocessing was applied as described in [4], including eddy current correction using FSL and fiber tracking by MedINRIA. The non-imaging dataset encompassed comprehensive patient information including demographic information (e.g., age, sex, education level), biomarker measurements (e.g., APOE-4), cognitive assessments (e.g., MMSE, CDR), and unstructured doctors' clinical notes. These components were integrated to generate comprehensive clinical narratives for each patient.

To evaluate the performance of our proposed method, we analyzed the most important image-text token pairs from different perspectives. First, we identified the key disease-related brain sub-networks and their corresponding clinical report tokens. Then, we highlighted the most significant disease-related text tokens and their associated brain sub-networks. The model was implemented using a vanilla ViT architecture comprising 4 layers with 256-dimensional embeddings. It was trained for 32 epochs on a single NVIDIA A6000 GPU with a batch size of 8, using the AdamW optimizer (learning rate = 1e-5). Data was split 80/20 for training/testing. Code is publicly available.

### 3.2 Critical brain sub-networks

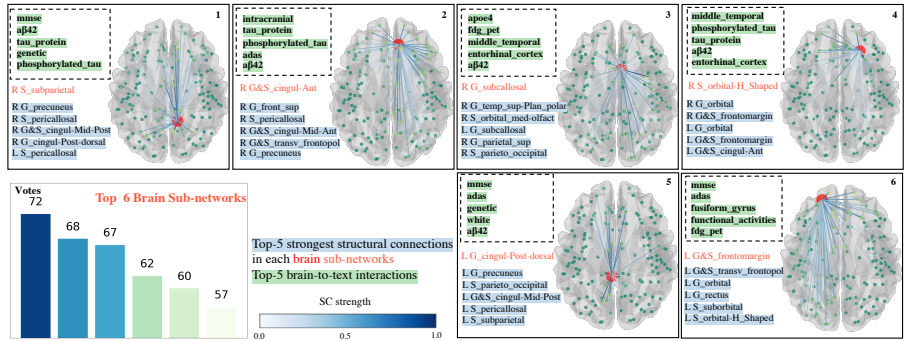

**Fig. 3.** Visualization of the Top-6 brain sub-networks and their associated key Brain-to-Text interactions. For each brain sub-network, we display its top-5 strongest structural connections and the top-5 most relevant tokens from clinical notes. The bottom-left presents voting results.

Fig. 3 visualizes the top 6 critical MCI-related brain sub-networks along with the most relevant text tokens derived from clinical report. For each sub-network, we highlight the top-5 significant Brain-to-Text interactions based on the cross-modality attention scores at connectome level. As expected, beyond MMSE and CDR scores, our model effectively identifies MCI-associated text tokens linked to specific sub-networks. Prior neuroimaging studies have demonstrated that MCI patients exhibit grey matter loss primarily in the parietal and posterior cingulate cortex, accompanied by white matter damage in these regions [5-8]. Our approach successfully captures these associations from both imaging and clinical text data, as reflected in panels 1 and 2 of Fig. 3.

More importantly, our model offers a powerful framework for improving our understanding of AD/MCI by linking multimodal data in practical clinical settings. previous studies have reported a puzzling discrepancy exists in AD research: structural atrophy primarily affects the medial temporal lobe, whereas metabolic dysfunction is more pronounced in the posterior association cortices [5]. This misalignment suggests that metabolic alterations may precede structural degeneration during AD progression. Our findings align with this hypothesis, as seen in panels 1 of Fig. 3, where the brain sub-network of the left and right *S_subparietal* regions shows a strong association with *tau*-related tokens. This connection is particularly significant because it not only reflects brain metabolic patterns but also corresponds to the fundamental pathological mechanisms of AD - the accumulation of extracellular amyloid plaques and intracellular neurofibrillary tangles (tau pathology). These pathological changes ultimately drive neuronal loss, cortical atrophy, and cognitive decline.

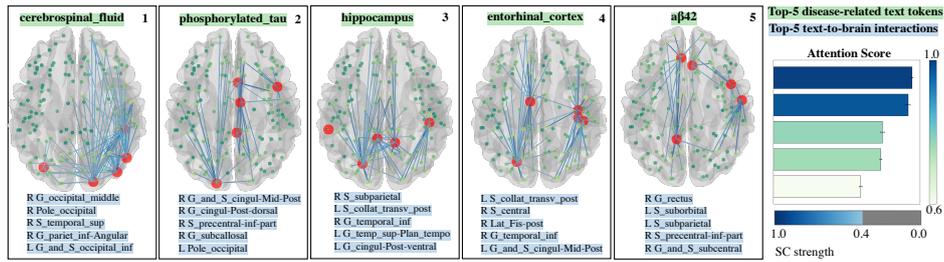

**Fig. 4.** Visualization of Top 5 disease-related text tokens and their associated key Text-to-Brain interactions. Each displaying a significant token and its corresponding top-5 strongly associated brain sub-networks. The bottom-right panel presents attention scores comparing MCI and NC groups, identifying tokens from the clinical reports that are most influential for disease diagnosis.

**3.3 Key clinical text tokens.** Fig. 4 presents the top-5 MCI-related text tokens extracted from the clinical reports and their corresponding significant Text-to-Brain interactions. To analyze these associations, we first identify the most influential tokens in diagnostic performance and then determine the top-5 most relevant brain sub-networks linked to each token. A notable example is the strong association between the token *phosphorylated_tau* and cingulate brain regions as shown in panel 2 of Fig. 4. Previous studies have demonstrated that the degree of tau protein deposition in these regions is closely correlated with cognitive decline, as evidenced by MRI and PET analyses [9-13]. Our model successfully captures this relationship, identifying brain sub-networks *G_and_S_cingul-Mid-Post* and *G_cingul-Post-dorsal* through text analysis, which closely aligns with established neuroimaging findings. This finding further validates our model's capability to uncover meaningful multimodal relationships between clinical text and neuroimaging data.

**3.4. Evaluation of MCI/NC prediction performance.** Our model is trained for the MCI/NC classification task and we compare its performance with the most recent approaches. For a fair comparison, Table 1 summarizes the overall prediction performance of recent MCI studies across three categories. As shown in Table 1, multi-imaging approaches generally outperform single-modal methods, highlighting the complementary insights different imaging modalities provide in characterizing MCI. However, while clinical reports contain critical first-hand information essential for diagnosis, few studies have explored the predictive potential of integrating text-based clinical reports with neuroimaging data. In this study, we combine DTI-derived brain connectome with clinical reports to enhance MCI prediction. Our experimental results demonstrate that our multimodal approach achieves the highest prediction accuracy compared to both single- and multi- imaging methods, underscoring the advantage of incorporating clinical text data in MCI prediction.

**3.5. Ablation study.** First, we assess the contribution of each modality - imaging and text - to the prediction task. As shown in Table 2 (A), both brain SC and clinical reports play a significant role in MCI prediction. Second, we evaluate the impact of different cross-modality alignment strategies at both subject and connectome levels, as summarized in Table 2 (B). The results indicate that integrating both modalities provides complementary benefits, leading to improved overall prediction performance. Moreover, the connectome-level alignment strategy not only enhances diagnostic accuracy but also uncovers meaningful interactions between brain sub-networks and corresponding clinical report tokens. These findings comprehensively validate the effectiveness of our proposed model and highlight its potential for advancing multimodal MCI diagnosis.

Table 1. Comparison of identification performance with other recent state-of-the-art methods. ACC: Accuracy; SEN: Sensitivity; SPE: Specificity; F1: F1-score

| Method | Modality | NC/MCI | ACC% | SEN% | SPE% | F1% |
|---|---|---|---|---|---|---|
| *(A) Single-Imaging Modality Study* | | | | | | |
| Shao et al. (2020) [14] | PET | 160/460 | 65.61 | 79.19 | 48.01 | - |
| Shao et al. (2020) [14] | MRI | 225/120 | 69.14 | 79.34 | 55.92 | - |
| Song et al. (2024) [15] | DTI | 27/10 | - | 70.0 | 85.2 | - |
| Zhang et al. (2025) [16] | rsfMRI | 225/120 | 89.1 | 91.4 | 97.2 | 90/3 |
| *(B) Multi-Imaging Modality Study* | | | | | | |
| Zhou et al. (2021) [17] | MRI+PET | 142/82 | 88.25 | 79.74 | 91.58 | - |
| Shi et al. (2022) [18] | MRI+PET | 52/99 | 80.73 | 85.98 | 70.90 | 85.3 |
| Zhang et al. (2021) [19] | DTI+rsfMRI | 116/98 | 92.70 | - | 89.90 | - |
| Zhang et al. (2023) [20] | DTI+rsfMRI | 116/93 | 92.30 | 83.22 | 89.99 | - |
| *(C) Imaging-Non-Imaging Modality Study* | | | | | | |
| Janani et al. (2021) [21] | MRI+record | 132/104 | 84.0 | - | 78.5 | 77.0 |
| Xue et al. (2024) [22] | MRI+record | 868/1119 | 59.5 | 55.0 | 64.0 | 41.0 |
| Zhang et al. (2025) [23] | MRI+text | 5025/7111 | - | 91.5 | - | 89.0 |
| Song et al. (2024) [16] | DTI+record | 27/10 | - | 90.0 | 81.5 | - |
| **Ours** | DTI+text | 301/117 | **92.94** | **93.0** | 85.16 | **93.0** |

Table 2. Quantitative ablation results: I-only and C-only means using imaging and clinical texts as input respectively; $\mathcal{L}_{cl}$ and $\mathcal{L}_{sl}$ represent connectome- and subject- level alignment.

| (A) Modality Ablation | | | | | (B) Alignment Strategy Ablation | | | | |
|---|---|---|---|---|---|---|---|---|---|
| Method | ACC% | SEN% | SPE% | F1% | Method | ACC% | SEN% | SPE% | F1% |
| I-only | 73.77 | 74.70 | 72.11 | 73.38 | w/o $\mathcal{L}_{cl}, \mathcal{L}_{sl}$ | 90.59 | 88.40 | 91.15 | 89.75 |
| C-only | 86.96 | 87.70 | 86.98 | 87.34 | w/o $\mathcal{L}_{cl}$ | 91.76 | 92.0 | 82.76 | 92.0 |
| **Ours** | 92.94 | 93.0 | 85.16 | 93.0 | **Ours** | 92.94 | 93.0 | 85.16 | 93.0 |

## 4 Conclusion

We introduce a novel multimodal approach that bridges brain connectomes and clinical reports at both subject and connectome levels for early AD diagnosis, enhancing both accuracy and interpretability. Our findings reveal several valuable insights derived solely from DTI-formed brain network and text information without additional imaging data, highlighting the potential of clinical reports analysis in uncovering neurodegenerative disease patterns. In the future, we will further explore more imaging modalities for a more comprehensive and generalizable brain presentation, further advancing AI-assisted diagnosis in neurodegenerative research.

**Acknowledgments.** This work was supported by the National Institutes of Health (R01AG075582 and RF1NS128534)

**Disclosure of Interests.** The authors have no competing interests to declare that are relevant to the content of this article.

10      J. Zhang, X. Yu et al.8. Teipel S J, Meindl T, Wagner M, et al. Longitudinal changes in fiber tract integrity in healthy aging and mild cognitive impairment: a DTI follow-up study[J]. Journal of Alzheimer's Disease, 2010, 22(2): 507-522.
9. Bailly M, Destrieux C, Hommet C, et al. Precuneus and Cingulate Cortex Atrophy and Hypometabolism in Patients with Alzheimer's Disease and Mild Cognitive Impairment: MRI and 18F-FDG PET Quantitative Analysis Using FreeSurfer[J]. BioMed research international, 2015, 2015(1): 583931.
10. Putcha D, Eckbo R, Katsumi Y, et al. Tau and the fractionated default mode network in atypical Alzheimer's disease[J]. Brain Communications, 2022, 4(2): fcac055.
11. Lantero-Rodriguez J, Camporesi E, Montoliu-Gaya L, et al. Tau protein profiling in tauopathies: a human brain study[J]. Molecular Neurodegeneration, 2024, 19(1): 54.
12. Pezzoli S, Giorgio J, Chen X, et al. Cognitive aging outcomes are related to both tau pathology and maintenance of cingulate cortex structure[J]. Alzheimer's & Dementia, 2025: e14515.
13. Pezzoli S, Giorgio J, Martersteck A, et al. Successful cognitive aging is associated with thicker anterior cingulate cortex and lower tau deposition compared to typical aging[J]. Alzheimer's & Dementia, 2024, 20(1): 341-355.
14. Shao W, Peng Y, Zu C, et al. Hypergraph based multi-task feature selection for multimodal classification of Alzheimer's disease[J]. Computerized Medical Imaging and Graphics, 2020, 80: 101663.
15. Song Q, Peng J, Shu Z, et al. Predicting Alzheimer's progression in MCI: a DTI-based white matter network model[J]. BMC Medical Imaging, 2024, 24(1): 103.
16. Zhang J, Lyu Y, Yu X, et al. Classification of Mild Cognitive Impairment Based on Dynamic Functional Connectivity Using Spatio-Temporal Transformer[J]. arXiv preprint arXiv:2501.16409, 2025.
17. Zhou P, Jiang S, Yu L, et al. Use of a sparse-response deep belief network and extreme learning machine to discriminate Alzheimer's disease, mild cognitive impairment, and normal controls based on amyloid PET/MRI images[J]. Frontiers in Medicine, 2021, 7: 621204.
18. Shi Y, Zu C, Hong M, et al. ASMFS: Adaptive-similarity-based multi-modality feature selection for classification of Alzheimer's disease[J]. Pattern Recognition, 2022, 126: 108566.
19. Zhang L, Wang L, Gao J, et al. Deep fusion of brain structure-function in mild cognitive impairment[J]. Medical image analysis, 2021, 72: 102082.
20. Zhang L, Na S, Liu T, et al. Multimodal deep fusion in hyperbolic space for mild cognitive impairment study[C]//International Conference on Medical Image Computing and Computer-Assisted Intervention. Cham: Springer Nature Switzerland, 2023: 674-684.
21. Venugopalan J, Tong L, Hassanzadeh H R, et al. Multimodal deep learning models for early detection of Alzheimer's disease stage[J]. Scientific reports, 2021, 11(1): 3254.
22. Xue C, Kowshik S S, Lteif D, et al. AI-based differential diagnosis of dementia etiologies on multimodal data[J]. Nature Medicine, 2024, 30(10): 2977-2989.
23. Zhang J, Yu X, Lyu Y, et al. Brain-Adapter: Enhancing Neurological Disorder Analysis with Adapter-Tuning Multimodal Large Language Models[J]. arXiv preprint arXiv:2501.16282, 2025.